\documentclass{article}

\usepackage{arxiv}

\usepackage[utf8]{inputenc} 
\usepackage[T1]{fontenc}    
\usepackage{hyperref}       
\usepackage{url}            
\usepackage{booktabs}       
\usepackage{amsfonts}       
\usepackage{nicefrac}       
\usepackage{microtype}      
\usepackage{lipsum}		
\usepackage{graphicx}
\usepackage{natbib}
\usepackage{doi}
\usepackage{amsmath}
\usepackage{multirow}
\usepackage{wrapfig}
\usepackage{makecell}

\newcommand{\sn}[2]{\ensuremath{{#1}{\times} 10^{#2}}}
\newcommand{\tn}[0]{\textsuperscript{\textdagger}}

\title{Multi-task problems are not multi-objective}

\author{ Michael Ruchte
    \\
	Representation Learning Lab\\
	University of Freiburg\\
	Freiburg, Germany \\
	\texttt{ruchtem@cs.uni-freiburg.de} \\
	\And
	Josif Grabocka \\
	Representation Learning Lab\\
	University of Freiburg\\
	Freiburg, Germany \\
	\texttt{grabocka@cs.uni-freiburg.de} \\
}

\date{}

\hypersetup{
pdftitle={Multi-task problems are not multi-objective},
pdfsubject={Machine Learning},
pdfauthor={Michael Ruchte, Josif Grabocka},
pdfkeywords={Machine Learning, Multi-Objective Optimization},
}

\begin{document}
\maketitle

\begin{abstract}
	Multi-objective optimization (MOO) aims at finding a set of optimal configurations for a given set of objectives. A recent line of work applies MOO methods to the typical Machine Learning (ML) setting, which becomes multi-objective if a model should optimize more than one objective, for instance in fair machine learning. These works also use Multi-Task Learning (MTL) problems to benchmark MOO algorithms treating each task as independent objective.
	
	In this work we show that MTL problems do not resemble the characteristics of MOO problems. In particular, MTL losses are not competing in case of a sufficiently expressive single model. As a consequence, a single model can perform just as well as optimizing all objectives with independent models, rendering MOO inapplicable. We provide evidence with extensive experiments on the widely used Multi-Fashion-MNIST datasets. Our results call for new benchmarks to evaluate MOO algorithms for ML. Our code is available at: \url{https://github.com/ruchtem/moo-mtl}.
\end{abstract}


\section{Introduction}

Multi-Objective Optimization (MOO) is a large field with applications ranging from car design to Adaptive Bitrate Control \citep{moo-benchmarks, vehicle-design, adaptive-bitrate}. It is defined as finding the optimal trade-offs between $J$ unweighted objectives:\footnote{In this work we ignore further information like constraints on $x$ or $f_j(x)$ or some predefined ordering of objectives.}

\begin{align}
    \label{eq:moo_definition}
    \min_{x \in \mathcal{X}} \left( f_1(x), f_2(x), \hdots, f_J(x) \right)
\end{align}

With recent advances in gradient-based MOO \citep{fliege, desideri} an emerging line of deep learning research focuses on applying MOO algorithms for deep neural networks \citep{mgda, pmtl, phn}. These works include casting Multi-Task Learning (MTL) problems into the MOO setting or balancing accuracy and fairness thereby optimizing neural networks for more than one loss. Different to fair machine learning, MTL problems come with the availability of several large-scale datasets and advanced tasks and thus is often used to benchmark MOO approaches \citep{pmtl, phn, cosmos}.

In this work we demonstrate that MTL problems differ from MOO problems in a significant way. In particular, the objectives are not necessarily competitive as the network has dedicated parameters for every objective. As a result we can optimize all objectives with just one model (one solution) rather than trading off different objectives which renders MOO inapplicable. In fact, we show that competition is caused by limited model capacity offering the simple solution to just increase the parameters to solve MOO problems.

To confirm our claim we make extensive experiments on the frequently used Multi-MNIST datasets \citep{pmtl}.

\section{Related work}
\label{sec:related-work}

The foundations for gradient-based MOO were created by \citep{fliege} and \citep{desideri}. Particularly, \citet{desideri} derived the Multiple Gradient Descent Algorithm (MGDA) which makes a step towards the minimum norm of the gradients with respect to each objective function, guaranteeing a Pareto-optimal solution. \citet{mgda} applied this to MTL problems and provided an approximate solution for neural networks which come with a large parameter space. As they demonstrate, applying MOO methods to fit a network for multiple tasks can be a great advantage. This is not to be confused with our claim which questions benchmarking MOO algorithms on MTL problems.

The first work to generate a full Pareto front was \cite{pmtl}. They proposed to train $k$ models independently, each bound to a region of the objective space, naming their method Pareto Multi-Task Learning (PMTL). The regions are defined by preference rays which reflect a practitioner's trade-off choices. Two works use hypernetworks \citep{hypernetworks} to condition the parameters on any possible preference ray \citep{phn, cpmtl}. This way they can approximate the full Pareto front at inference time. We consider Pareto Hypernetworks (PHN) \citep{phn} as baseline as its code is available. \citet{cosmos} follow a similar idea but condition the model in input feature space by concatenating randomly sampled preference rays to the input features and add a penalty to ensure a well-spread Pareto front.

All methods rely strongly on MTL problems to demonstrate their effectiveness, mainly:   Multi-Fashion-MNIST \citep{pmtl}, NYUv2 \citep{nyuv2}, and CityScapes \citep{cityscapes}. \citet{phn} and \citet{cosmos} additionally evaluate on fairness problems (e.g. Adult \citep{adult}). 

\section{Multi-task problems are not multi-objective}

In this section we compare the MTL setting with original MOO setting. For MTL the model parameters $\theta$ depend not only on the objective functions $\mathcal{L}$ but also on mini-batches of training data $D$. Thus with a slight change towards common machine learning notation Equation~\ref{eq:moo_definition} becomes

\begin{align}
    \label{eq:dlmoo_definition}
    \min_{\theta \in \Theta} \left( \mathcal{L}_1(\theta | D), \mathcal{L}_2(\theta | D), \hdots, \mathcal{L}_J(\theta | D) \right)
\end{align}

This renders MOO a stochastic problem with (generally) non-convex loss functions resulting in approximate solutions with many different local optima. Maybe somewhat more challenging but not fundamentally different to Equation~\ref{eq:moo_definition}. This is the problem to be solved in e.g. fair machine learning where a model should be both accurate but also non-discriminative with respect to some attributes like race or gender. Another example would be recommender systems which should optimize both customer experience and revenue \citep{mgda-recommender-large}.

For MTL, however, each loss is independent of the others as each loss is tied to a specific output of the model. Hence a perfectly valid though naive approach is to optimize a different model for each output (Equation~\ref{eq:mtl_definition}).

\begin{align}
    \label{eq:mtl_definition}
    \min_{\theta_1 \in \Theta_1, \theta_2 \in \Theta_2, \hdots, \theta_J \in \Theta_J} \left( \mathcal{L}_1(\theta_1 | D), \mathcal{L}_2(\theta_2 | D), \hdots, \mathcal{L}_J(\theta_J | D)  \right)
\end{align}

For efficiency reasons, however, it is common to share some parts of the parameters across tasks \citep{mtl}:

\begin{align}
    \label{eq:mtlmoo_definition}
    \min_{\theta_s \in \Theta_s, \theta_1 \in \Theta_1, \hdots, \theta_J \in \Theta_J} (\mathcal{L}_1(\theta_s, \theta_1 | D), \mathcal{L}_2(\theta_s, \theta_2 | D), \hdots, \mathcal{L}_J(\theta_s, \theta_J | D) )
\end{align}

where $\theta_s$ are shared parameters across all tasks and $\theta_j$ are similar to Equation~\ref{eq:mtl_definition} task-specific parameters. This is different to Equation~\ref{eq:moo_definition} and Equation~\ref{eq:dlmoo_definition}, where there exists no objective-specific part of $x$ or $\theta$. With enough capacity in either $\theta_s$ or $\theta_j$, the model should be able to optimize all objectives simultaneously as in Equation~\ref{eq:mtl_definition}.

\section{Experiments}

We consider training on a single task (\textit{Single Task} method) to obtain the optimal score on any given MTL problem (i.e. Equation~\ref{eq:mtl_definition}). Additionally, we benchmark six methods: \textit{Uniform} which minimizes the uniformly weighted sum of all objectives $\min \sum^J_{j=1} \frac{1}{J} \mathcal{L}_j$; \textit{MGDA} \citep{mgda}; \textit{PMTL} \citep{pmtl}; \textit{PHN} (linear scalarization if not mentioned otherwise) \citep{phn}; and \textit{COSMOS} \citep{cosmos}.

\subsection{Experimental setup}

We follow the experimental setup of the baselines with fixed 100 training epochs, 256 batch size, and Adam \citep{adam}. As reference point for Hypervolume (HV) \citep{hypervolume} we use $(1, 1)$. As model we use LeNet following \citet{pmtl}. We use the Multi-Fashion-MNIST dataset \citep{pmtl} with predefined splits: 108k training, 12k validation, and 20k test images. In these datasets two instances are merged into one image with a large overlap. The two tasks are to predict the class of each instance. We use the validation images only for HPO and do not merge them for final training. We report final results averaged over ten random seeds. We perform all experiments on Nividia GeForce RTX 2080 GPUs.

\begin{wraptable}{R}{5.7cm}
\vspace{-.35cm}
\centering
\begin{tabular}{lc}\toprule
\bf Parameter      & \bf Range \\ \midrule
Learning rate           & \makecell{$\sn{1}{-2}, \sn{7.5}{-3}$,\\$\sn{5}{-3}, \sn{2.5}{-3}$,\\$ \hdots, \sn{1}{-4}$} \\
Weight decay            & \makecell{$\sn{1}{-1}, \sn{7.5}{-2}$,\\$\sn{5}{-2}, \sn{2.5}{-2}$,\\$\hdots, \sn{1}{-4}$} \\
\makecell{Learning rate\\scheduler} & \makecell{Cosine Annealing,\\Step-wise, None} \\ \bottomrule

\end{tabular}
\caption{Method-independent hyperparameter search spaces}
\label{tab:hp_search_spaces}
\vspace{-.4cm}
\end{wraptable}

\paragraph{Hyperparameter optimization.} We perform random search over the hyperparameter search spaces defined in Table~\ref{tab:hp_search_spaces}. Learning rate scheduler specifics are  $\sn{1}{-6}$ as minimum learning rate for Cosine Annealing \citep{cosineannealing} and Step-wise with milestones at epoch 33 and 66 with multiplier 0.1. We sample 100 configurations uniformly at random, which covers approximately 30\% of the $9 \times 13 \times 3 = 351$ possible combinations. We select the best configuration based on Hypervolume based on cross-entropy loss measured on the validation set.

Some methods come with their own set of hyperparameters. For MGDA this is gradient normalization $\in \{\text{l2}, \text{loss}, \text{loss+}, \text{None} \}$. For PHN this is the internal solver $\in \{ \text{ls}, \text{EPO} \}$ and the Dirichlet sampling parameter $\alpha \in \{0.1, 0.2, 0.5, 1, 1.2, 1.5\}$. For COSMOS this is $\alpha$ and the cosine penalty weight $\lambda \in \{1, 2, 4, 8, 16 \}$. The ranges are chosen to cover the original values plus one value beyond. Method-specific HPO covers a smaller part of the search space, approximately 7, 2.37, and 0.95 Percent for MGDA, PHN, and COSMOS, respectively. We denote these methods with an asterix ($^*$). For PHN and COSMOS we provide also results not optimizing the internal solver, $\alpha$, or $\lambda$ taking the original values reported by the authors. See Section~\ref{sec:hpo_results} in the supplementary for the parameters found by HPO.

\subsection{Results}
\label{sec:results}

\begin{figure}
    \centering
    \includegraphics[width=\textwidth]{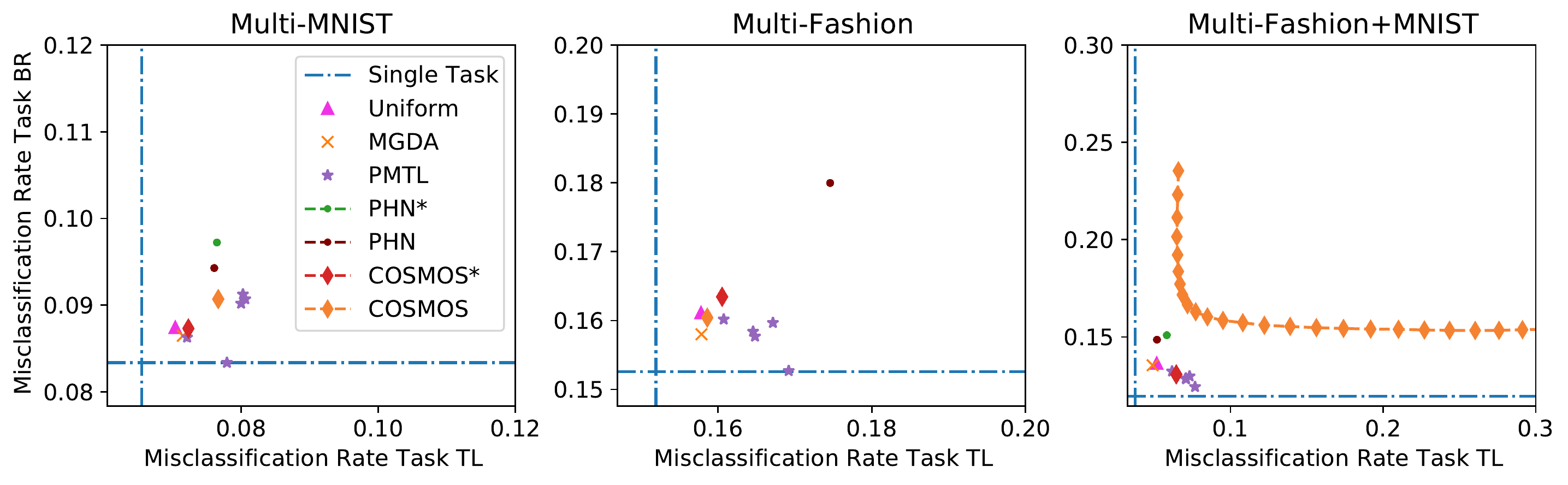}
    \caption{Results on Multi-MNIST dataset for different methods. Linear scalarization with uniform weights (Uniform) is a competetive baseline despite its simplicity given its hyperparameters are properly tuned.}
    \label{fig:all_methods_mcr}
\end{figure}

\begin{table}
    \centering
    \begin{tabular}{lccccccc} \toprule
\multirow{2}{*}{\bf Method}  & \multirow{2}{*}{\bf Params}  & \multicolumn{2}{c}{\bf Multi MNIST} & \multicolumn{2}{c}{\bf Multi Fashion} & \multicolumn{2}{c}{\bf multi fashion+mnist} \\ \cmidrule(lr){3-4} \cmidrule(lr){5-6} \cmidrule(lr){7-8}
                         &        & HV $\uparrow$ & $\Delta$ ST $\downarrow$ & HV $\uparrow$ & $\Delta$ ST $\downarrow$ & HV $\uparrow$ & $\Delta$ ST $\downarrow$ \\ \midrule
PMTL                     &  211k  & \it0.8509 $\pm$ 0.0023 & \it0.0058 & \it0.7139 $\pm$ 0.0056 & \it0.0048 & \it0.8220 $\pm$ 0.0062 & \it0.0254 \\
PHN$^*$                  &  \multirow{2}{*}{3,243k} & 0.8337 $\pm$ 0.0041 & 0.0229 & 0.6769 $\pm$ 0.0143 & 0.0417 & 0.7996 $\pm$ 0.0053 & 0.0478 \\
PHN                      &        & 0.8368 $\pm$ 0.0046 & 0.0198 & 0.6769 $\pm$ 0.0143 & 0.0417 & 0.8073 $\pm$ 0.0051 & 0.0401 \\
COSMOS$^*$ & \multirow{2}{*}{42k}    & \it0.8467 $\pm$ 0.0039 & \it0.0099 & 0.7023 $\pm$ 0.0087 & 0.0164 & 0.8132 $\pm$ 0.0026 & 0.0342 \\ 
COSMOS                   &     & 0.8396 $\pm$ 0.0047 & 0.0170 & 0.7064 $\pm$ 0.0055 & 0.0123 & 0.7915 $\pm$ 0.0111 & 0.0559 \\ \midrule
MGDA$^*$                 &  42k   & \it0.8482 $\pm$ 0.0060 & \it0.0084 & \it0.7091 $\pm$ 0.0040 & \it0.0096 & \it0.8222 $\pm$ 0.0046 & \it0.0252 \\
Uniform                  &  42k  & \it0.8484 $\pm$ 0.0091 & \it0.0083 & \it0.7065 $\pm$ 0.0111 & \it0.0121 & \it0.8189 $\pm$ 0.0067 & \it0.0285 \\ \midrule \midrule
Single Task              &  42k  & 0.8566 $\pm$ 0.0023 & - & 0.7187 $\pm$ 0.0036 & - & 0.8474 $\pm$ 0.0028 & - \\ \bottomrule
    \end{tabular}
    \vspace{0.2cm}
    \caption{Hypervolume (HV) based on misclassification rates of properly optimized methods for the Multi-Mnist datasets (larger is better). $\Delta$ ST denotes the distance of each method to Single Task, an indication whether the problem is multi-objective. $^*$ denotes optimizing method-specific hyperparameters as well. \textit{Italic} denotes the second-best method after Single Task. Linear scalarization with uniform weights (Uniform) is a competetive baseline despite its simplicity given its hyperparameters are properly tuned. Surprisingly, almost all Pareto front generating models (upper section) are not competitive even though their hyperparameters are also tuned to maximize HV.}
    \label{tab:results_mcr}
\end{table}

We show the results in terms of Misclassification Rate (MCR) in Figure~\ref{fig:all_methods_mcr} and Table~\ref{tab:results_mcr}. We can see that Uniform despite being by far the most simple method is competitive compared to the more complex methods and achieves nearly the same performance as Single Task. The gap between Uniform and Single Task is larger for Multi-Fashion indicating that here the two tasks are indeed competing. We hypothesis, however, this is due to the limited capacity of the model and results with different model size in Section~\ref{sec:ablations}. For scores on each task and cross-entropy results see Section~\ref{sec:more_results} in the supplementary.

Unsurprisingly, using HPO improves performance of all methods when comparing to the numbers reported by the authors. For instance, PMTL reports Single Task MCR on Multi-MNIST Top-Left of 0.91 and Bottom-Right of 0.88, whereas we get 0.93 and 0.91, respectively. Same is true for the other datasets and methods. This underlines the importance of HPO for proper method evaluation. Particularly, surpassing Single Task models in MTL settings should raise concerns regarding overfitting and thus unfair comparisons.

Somewhat surprising is that even methods which can generate a Pareto front (especially PHN and COSMOS), do not produce one when their learning rate, weight decay and scheduler are optimized for validation HV. This problem was already mentioned by \citet{cosmos} and their main motivation to introduce the cosine penalty. We double checked and there were HPO configurations sampled which resulted in a Pareto front but actually achieved a much lower HV score. We see this as a further indication that we are not looking at an MOO problem.

\subsection{Ablations}
\label{sec:ablations}

\begin{wraptable}{R}{7.1cm}
\vspace{-0.35cm}
\centering
\begin{tabular}{lcc}\toprule
Dataset     & Single Task HV  & Uniform HV \\ \midrule
MM   & 0.8551 $\pm$ 0.0033 & 0.8488 $\pm$ 0.0056 \\
MF   & 0.7234 $\pm$ 0.0028 & 0.7122 $\pm$ 0.0075 \\
MFM  & 0.8442 $\pm$ 0.0034 & 0.8200 $\pm$ 0.0068 \\ \bottomrule
\end{tabular}
\caption{Grid seach HV results.}
\label{tab:results_grid_search}
\end{wraptable}

\paragraph{Grid search.}
To quantify a possible negative impact of random search in HPO we also perform grid search over all possible 351 hyperparameter combinations and report the results for a subset of the methods in Table~\ref{tab:results_grid_search}. Datasets are abbreviated as MM (Multi-MNIST), MF (Multi-Fashion), and MFM (Multi-Fashion-MNIST). There are almost no significant differences when comparing to Table~\ref{tab:results_mcr} which demonstrates that randomly searching about one third of the search space is a valid approach in our case.

\begin{figure}
    \centering
    \includegraphics[width=\textwidth]{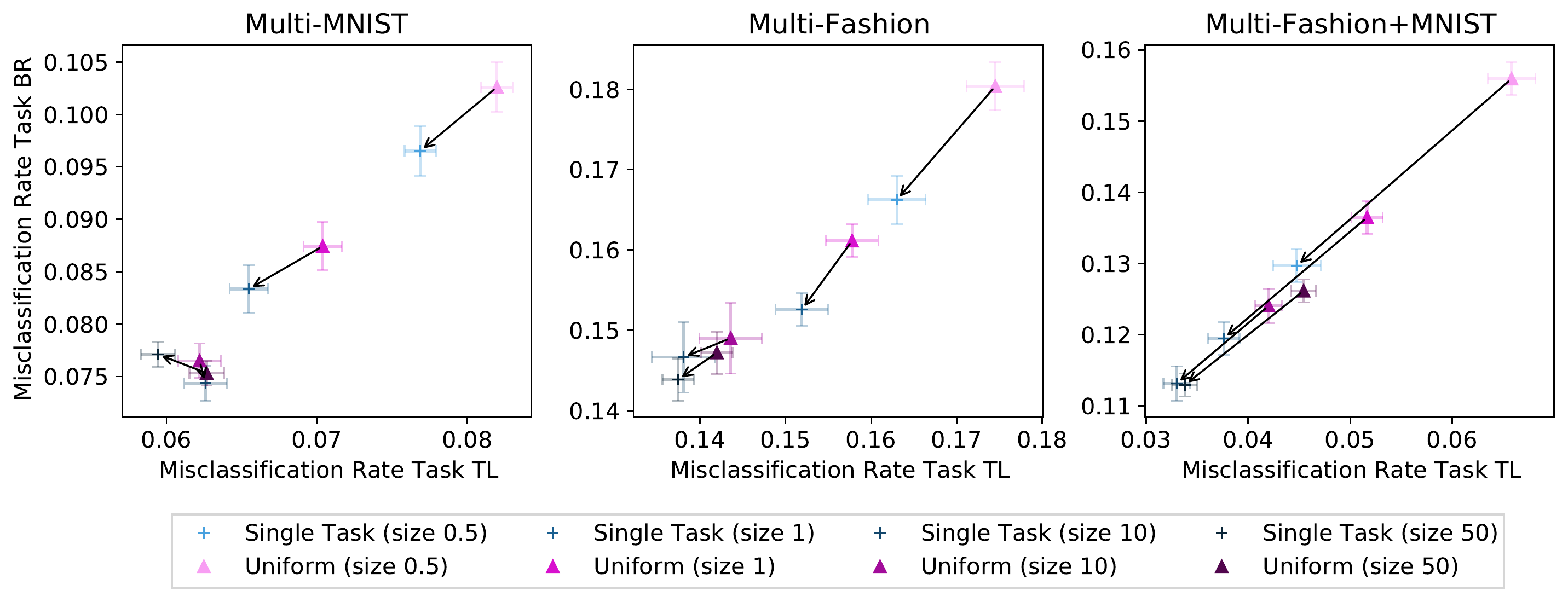}
    \caption{Results for Single Task and Uniform when varying the capacity of the model. Errorbars denote standard deviation. We can see that the distance between Single Task and Uniform (black arrows) as well as the absolute error decrease significantly when increasing the capacity of the model.}
    \label{fig:capaciy_ablation}
\end{figure}

\paragraph{Model capacity.}
We ask the question whether capacity plays a crucial role in rendering MTL problems multi-objective with tasks competing against each other. For this we alter the capacity of the model by changing the number of channels with a multiplier. Specifically, we change the model to have $10c$ channels in the first layer and $20c$ channels in the second layer where $c \in \{0.5, 1, 10, 50 \}$ is the channel multiplier (i.e. we only alter $\theta_s$).

The results are presented in Figure~\ref{fig:capaciy_ablation} and Table~\ref{tab:capacity_ablation}. We can see that the distance between Single Task and Uniform scores indeed depends on the capacity of the models. Reducing capacity increases the distance, increasing capacity decreases the distance up to a saturation point where error rates almost match. Interestingly, for Multi-Fashion-MNIST there is still a considerable gap between single task and uniform performance. This could be an indication that not only capacity matters but also the similarity of the tasks.

The experiments support the hypothesis that for MTL problems it is more beneficial to increase the capacity instead of treating it as a MOO problem trying to balance presumed conflicts between losses. This is especially true for methods which inherently increase the capacity as PHN does. Multiplying the channels by 10 and using uniform scalarization yields a better Hypervolume with about 25\% of the parameters compared to PHN (0.03 HV improve on Multi-MNIST). 

\begin{table}
\centering
\footnotesize
\begin{tabular}{lcc cccccc}\toprule
\multirow{2}{*}{\bf Method}      & \multirow{2}{*}{\bf $c$} & \multirow{2}{*}{\bf Params} & \multicolumn{2}{c}{\bf Multi Mnist} & \multicolumn{2}{c}{\bf Multi fashion}  & \multicolumn{2}{c}{\bf Multi fashion+mnist} \\ \cmidrule(lr){4-5} \cmidrule(lr){6-7} \cmidrule(lr){8-9}
            & &  & HV     & $\Delta$ ST & HV     & $\Delta$ ST & HV     & $\Delta$ ST  \\ \midrule
Single Task & \multirow{2}{*}{0.5x} & \multirow{2}{*}{20k} & 0.8340 $\pm$ 0.0028 & - & 0.6978 $\pm$ 0.0045 & - & 0.8313 $\pm$ 0.0025 & -  \\
Uniform     &                       &                     & 0.8238 $\pm$ 0.0065 & 0.0102 & 0.6767 $\pm$ 0.0143 & 0.0212 & 0.7885 $\pm$ 0.0095 & 0.0428 \\ \midrule
Single Task & \multirow{2}{*}{1x} & \multirow{2}{*}{42k}   & 0.8566 $\pm$ 0.0023 & - & 0.7187 $\pm$ 0.0036 &- & 0.8474 $\pm$ 0.0028 & -  \\
Uniform     &                     &                       & 0.8484 $\pm$ 0.0091 & 0.0083 & 0.7065 $\pm$ 0.0111 & 0.0121 & 0.8189 $\pm$ 0.0067 & 0.0285 \\ \midrule
Single Task & \multirow{2}{*}{10x} & \multirow{2}{*}{863k} & 0.8677 $\pm$ 0.0022 & - & 0.7355 $\pm$ 0.0042 & - & 0.8576 $\pm$ 0.0025 & - \\
Uniform     &                      &                      & 0.8661 $\pm$ 0.0023 & 0.0016 & 0.7288 $\pm$ 0.0107 & 0.0067 & 0.8391 $\pm$ 0.0026 & 0.0184 \\ \midrule
Single Task & \multirow{2}{*}{50x} & \multirow{2}{*}{14,315k} & 0.8680 $\pm$ 0.0012 & - & 0.7384 $\pm$ 0.0033 & - & 0.8571 $\pm$ 0.0023 & -  \\
Uniform     &                      &                         & 0.8667 $\pm$ 0.0032 & 0.0013 & 0.7317 $\pm$ 0.0076 & 0.0067 & 0.8341 $\pm$ 0.0042 & 0.0230  \\ \bottomrule
\end{tabular}
\vspace{0.2cm}
\caption{Ablation results for different model sizes. $c$ denotes the channel multiplier, $\Delta$ ST denotes the distance of each method to Single Task. The results show that $\Delta$ ST decreases on overall when increasing the model capacity.}
\label{tab:capacity_ablation}
\vspace{-.2cm}
\end{table}

\section{Limitations}

One limitation is that we did not do proper HPO for methods with additional hyperparameters as this would result in unbearable computing costs. For the same reason we also used only one seed for HPO, despite the results in Section~\ref{sec:ablations} being aware that the obtained scores are noisy.

Also due to computing costs we decided to not repeat the analysis with larger models on one of the larger benchmarks mentioned in Section~\ref{sec:related-work}. We, however, do not see a reason why the results should not generalize to larger problems. This could be a direction for future research, as well as an analysis on the effect of altering the capacity of the task-specific heads, i.e. $\theta_j$.

\section{Conclusion}

In this work we showed that multi-task problems do not necessarily resemble the characteristics of multi-objective problems. One important aspect in rendering MTL tasks multi-objective is the capacity of the model and likely the similarity between the tasks. The results call for a new set of benchmarks which are truly multi-objective. Pointers could be Fairness \citep{phn, cosmos} and recommender systems \citep{mgda-recommender-large, mgda-recommender-small}. Especially the latter come with large datasets and require large models which is more suited for deep neural networks. Another pointer is \citet{deist2021multi} who evaluate on medical image segmentation with possibly conflicting labeling.

It required roughly 3100 GPU hours to compute the results presented here, where 2780 are due to the ablations.

\section*{Acknowledgments}

Michael Ruchte would like to thank Samuel Müller for the helpful discussions.

\bibliographystyle{plainnat}
\bibliography{references}

\newpage
\appendix
\section*{Appendix}

\section{HPO results}
\label{sec:hpo_results}

We show the hyperparameter values found by hyperparameter optimization for Section~\ref{sec:results} in Table~\ref{tab:optimal_hp} and for Section~\ref{sec:ablations} in Table~\ref{tab:optimal_hp_ablations}.

\begin{table}[!htbp]
\scriptsize
    \centering
    \begin{tabular}{llccccccccc}  \toprule
Parameter & Dataset & ST TL & ST BR & Uniform & MGDA & PMTL & PHN & PHN$^*$ & COSMOS & COSMOS$^*$ \\ \midrule
\multirow{3}{*}{\shortstack[l]{Lerning\\rate}}
          & MM     & 0.0005 & 0.00075 & 0.0025  & 0.001  & 0.00025 & 0.00075 & 0.001   & 0.00075 & 0.001   \\
          & MF     & 0.0025 & 0.001   & 0.00075 & 0.0025 & 0.00075 & 0.00075 & 0.00075 & 0.001   & 0.00075 \\
          & MFM    & 0.0025 & 0.00075 & 0.00075 & 0.001  & 0.00075 & 0.0005  & 0.0025  & 0.0075  & 0.001   \\ \midrule
\multirow{3}{*}{\shortstack[l]{Weight\\decay}}
          & MM     & 0.05   & 0.05    & 0.0075 & 0.01    & 0.075   & 0.05    & 0.075   & 0.0075 & 0.0075 \\
          & MF     & 0.01   & 0.01    & 0.01   & 0.005   & 0.05    & 0.075   & 0.075   & 0.0075 & 0.0075 \\
          & MFM    & 0.0075 & 0.01    & 0.01   & 0.01    & 0.05    & 0.075   & 0.005   & 0.0005 & 0.01   \\ \midrule
\multirow{3}{*}{\shortstack[l]{Learning\\rate\\scheduler}}
          & MM     & CA & MS & CA & CA & CA & CA & CA & CA & CA \\
          & MF     & MS & CA & CA & CA & CA & CA & CA & CA & CA \\
          & MFM    & CA & MS & CA & CA & CA & CA & CA & CA & MS \\ \midrule
\multirow{3}{*}{\shortstack[l]{Normal-\\ization}}
          & MM     &-   &-   &-   & loss      &- &- &- &- &- \\
          & MF     &-   &-   &-   & loss+     &- &- &- &- &- \\
          & MFM    &-   &-   &-   & none      &- &- &- &- &- \\ \midrule
\multirow{3}{*}{\shortstack[l]{Internal\\solver}}
          & MM     &- &- &- &- &- & linear\tn & epo    &- &- \\
          & MF     &- &- &- &- &- & linear\tn & linear &- &- \\
          & MFM    &- &- &- &- &- & linear\tn & linear &- &- \\ \midrule
\multirow{3}{*}{\shortstack[l]{$\alpha$}}
          & MM     &- &- &- &- &- & 0.2\tn    & 0.1    & 1.2\tn & 0.5 \\
          & MF     &- &- &- &- &- & 0.2\tn    & 0.2    & 1.2\tn & 1.0 \\
          & MFM    &- &- &- &- &- & 0.2\tn    & 1.2    & 1.2\tn & 0.2 \\ \midrule
\multirow{3}{*}{\shortstack[l]{$\lambda$}}
          & MM     &- &- &- &- &- &- &- & 2\tn & 1 \\
          & MF     &- &- &- &- &- &- &- & 2\tn & 2 \\
          & MFM    &- &- &- &- &- &- &- & 8\tn & 1 \\ \bottomrule
    \end{tabular}
    \vspace{0.2cm}
    \caption{Hyperparameters found by HPO, \tn denotes the values from the original works, here just for reference. MM=Multi MNIST, MF=Multi Fashion, MFM=Multi Fashion+MNIST.}
    \label{tab:optimal_hp}
\end{table}

\begin{table}[!htbp]
\footnotesize
    \centering
    \begin{tabular}{ll|ccc|ccc|ccc}  \toprule
Parameter & Dataset & \multicolumn{3}{c}{Single Task TL} & \multicolumn{3}{c}{Single Task BR} & \multicolumn{3}{c}{Uniform}   \\ 
& & 0.5x & 10x & 50x & 0.5x & 10x & 50x & 0.5x & 10x & 50x \\ \midrule
\multirow{3}{*}{\shortstack[l]{Lerning\\rate}}
          & MM     & 0.001 & 0.0001  & 0.0001  & 0.00075 & 0.00025 & 0.00025 & 0.0005 & 0.0005  & 0.0001 \\
          & MF     & 0.001 & 0.00075 & 0.0005  & 0.0025  & 0.001   & 0.0005  & 0.0025 & 0.00075 & 0.0005 \\
          & MFM    & 0.001 & 0.0005  & 0.00075 & 0.0025  & 0.00075 & 0.00025 & 0.0025 & 0.0005  & 0.0005 \\ \midrule
\multirow{3}{*}{\shortstack[l]{Weight\\decay}}
          & MM     & 0.05   & 0.01   & 0.01    & 0.01    & 0.01    & 0.01    & 0.0075 & 0.0075  & 0.0075 \\
          & MF     & 0.0075 & 0.01   & 0.01    & 0.005   & 0.0075  & 0.01    & 0.005  & 0.0075  & 0.0075 \\
          & MFM    & 0.01   & 0.01   & 0.01    & 0.0075  & 0.01    & 0.01    & 0.0075 & 0.0075  & 0.005 \\ \midrule
\multirow{3}{*}{\shortstack[l]{Learning\\rate\\scheduler}}
          & MM     & MS & MS & MS & MS & MS  & CA & CA  & MS & MS \\
          & MF     & MS & MS & MS & MS & CA  & CA & CA  & CA & MS \\
          & MFM    & MS & MS & CA & MS & MA  & MS & CA  & MS & CA \\  \bottomrule
    \end{tabular}
    \vspace{0.2cm}
    \caption{Hyperparameters found by HPO for the models with different capacity. The second row in the header refers to the channel multiplier (see Section \ref{sec:ablations}). MM=Multi MNIST, MF=Multi Fashion, MFM=Multi Fashion+MNIST. Learning rate scheduler abbreviations: MS=Step-wise (multi step), CA=Cosine Annealing.}
    \label{tab:optimal_hp_ablations}
\end{table}

\section{More results}
\label{sec:more_results}

\begin{figure}[!htbp]
    \centering
    \includegraphics[width=\textwidth]{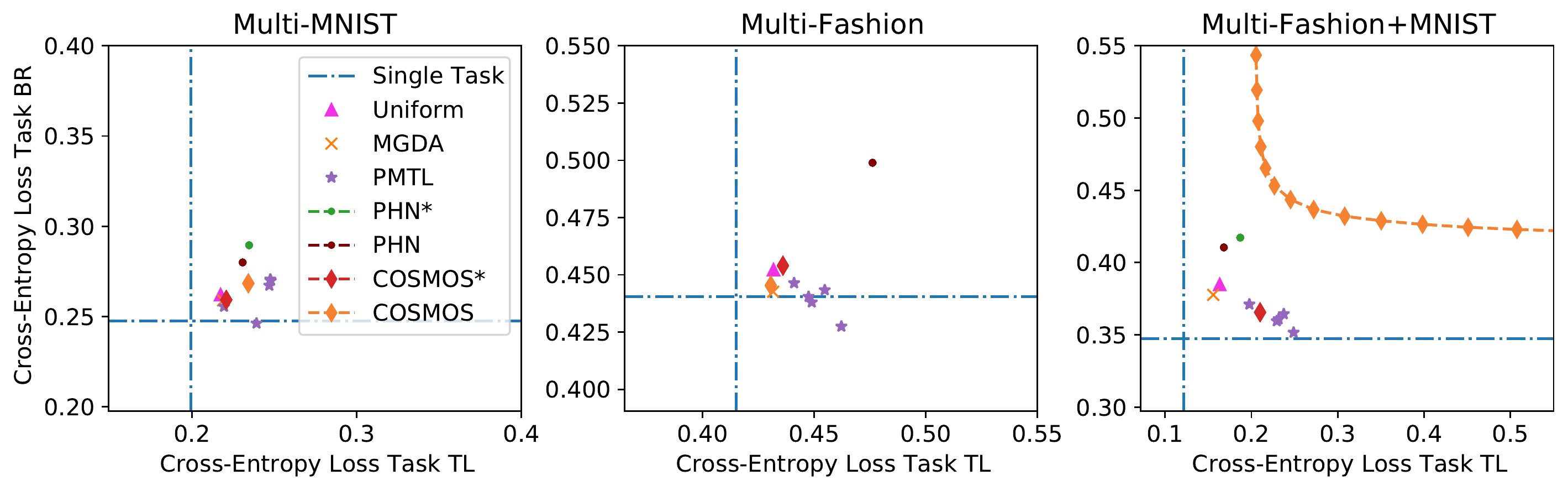}
    \caption{Results in terms of cross-entropy loss}
    \label{fig:all_methods_loss}
\end{figure}

We show the performance in terms of Misclassification rate for each task in Table~\ref{tab:results_individual_mcr}. The performance for each task in terms of cross entropy is shown in Figure~\ref{fig:all_methods_loss} and Table~\ref{tab:results_individual_loss}. An equivalent of Table~\ref{tab:results_mcr} for Hypervolume calculated based on cross entropy is presented in Table~\ref{tab:results_loss}.

\begin{table}[!htbp]
    \scriptsize
    \centering
    \begin{tabular}{lcccccc} \toprule
\multirow{2}{*}{\bf Method}  & \multicolumn{2}{c}{\bf Multi MNIST} & \multicolumn{2}{c}{\bf Multi Fashion} & \multicolumn{2}{c}{\bf multi fashion+mnist} \\ \cmidrule(lr){2-3} \cmidrule(lr){4-5} \cmidrule(lr){6-7}
             &  TL  & BR & TL  & BR & TL  & BR  \\ \midrule
Single Task & 0.0655 $\pm$ 0.0013 & 0.0834 $\pm$ 0.0023 & 0.1519 $\pm$ 0.0031 & 0.1526 $\pm$ 0.0020 & 0.0376 $\pm$ 0.0015 & 0.1195 $\pm$ 0.0023 \\
Uniform & 0.0704 $\pm$ 0.0051 & 0.0874 $\pm$ 0.0051 & 0.1578 $\pm$ 0.0070 & 0.1611 $\pm$ 0.0066 & 0.0517 $\pm$ 0.0041 & 0.1365 $\pm$ 0.0036 \\
MGDA & 0.0715 $\pm$ 0.0032 & 0.0865 $\pm$ 0.0036 & 0.1579 $\pm$ 0.0038 & 0.1580 $\pm$ 0.0030 & 0.0490 $\pm$ 0.0029 & 0.1354 $\pm$ 0.0035 \\ \midrule
ParetoMTL & 0.0806 $\pm$ 0.0030 & 0.0907 $\pm$ 0.0023 & 0.1646 $\pm$ 0.0064 & 0.1584 $\pm$ 0.0047 & 0.0732 $\pm$ 0.0044 & 0.1298 $\pm$ 0.0030 \\
PHN$^*$ & 0.0765 $\pm$ 0.0020 & 0.0972 $\pm$ 0.0031 & 0.1746 $\pm$ 0.0092 & 0.1800 $\pm$ 0.0085 & 0.0583 $\pm$ 0.0038 & 0.1509 $\pm$ 0.0036 \\
PHN & 0.0761 $\pm$ 0.0025 & 0.0943 $\pm$ 0.0031 & 0.1746 $\pm$ 0.0092 & 0.1800 $\pm$ 0.0085 & 0.0518 $\pm$ 0.0041 & 0.1486 $\pm$ 0.0030 \\
COSMOS$^*$ & 0.0723 $\pm$ 0.0022 & 0.0873 $\pm$ 0.0025 & 0.1605 $\pm$ 0.0051 & 0.1634 $\pm$ 0.0058 & 0.0645 $\pm$ 0.0012 & 0.1308 $\pm$ 0.0023 \\
COSMOS & 0.0767 $\pm$ 0.0028 & 0.0907 $\pm$ 0.0029 & 0.1586 $\pm$ 0.0026 & 0.1604 $\pm$ 0.0047 & 0.1081 $\pm$ 0.0069 & 0.1573 $\pm$ 0.0061 \\ \bottomrule
    \end{tabular}
    \vspace{0.2cm}
    \caption{Results for each task in terms of Misclassification rate (MRC)}
    \label{tab:results_individual_mcr}
\end{table}

\begin{table}[!htbp]
    \footnotesize
    \centering
    \begin{tabular}{lcccccc} \toprule
\multirow{2}{*}{\bf Method}  & \multicolumn{2}{c}{\bf Multi MNIST} & \multicolumn{2}{c}{\bf Multi Fashion} & \multicolumn{2}{c}{\bf multi fashion+mnist} \\ \cmidrule(lr){2-3} \cmidrule(lr){4-5} \cmidrule(lr){6-7}
                         & HV  & $\Delta$ ST & HV  & $\Delta$ ST & HV  & $\Delta$ ST \\ \midrule
Single Task & 0.6024 $\pm$ 0.0048 & 0      & 0.3272 $\pm$ 0.0043 & 0      & 0.5732 $\pm$ 0.0072 & 0      \\
Uniform     & 0.5777 $\pm$ 0.0191 & 0.0248 & 0.3116 $\pm$ 0.0202 & 0.0157 & 0.5147 $\pm$ 0.0148 & 0.0584 \\
MGDA$^*$    & 0.5789 $\pm$ 0.0158 & 0.0235 & 0.3168 $\pm$ 0.0072 & 0.0104 & 0.5254 $\pm$ 0.0102 & 0.0478 \\ \midrule
PMTL        & 0.5888 $\pm$ 0.0080 & 0.0136 & 0.3261 $\pm$ 0.0106 & 0.0011 & 0.5209 $\pm$ 0.0160 & 0.0523 \\
PHN$^*$     & 0.5438 $\pm$ 0.0097 & 0.0586 & 0.2631 $\pm$ 0.0247 & 0.0641 & 0.4738 $\pm$ 0.0119 & 0.0994 \\
PHN         & 0.5539 $\pm$ 0.0110 & 0.0485 & 0.2631 $\pm$ 0.0247 & 0.0641 & 0.4905 $\pm$ 0.0099 & 0.0827 \\
COSMOS$^*$  & 0.5772 $\pm$ 0.0102 & 0.0252 & 0.3082 $\pm$ 0.0172 & 0.0190 & 0.5011 $\pm$ 0.0069 & 0.0720 \\
COSMOS      & 0.5603 $\pm$ 0.0106 & 0.0421 & 0.3159 $\pm$ 0.0118 & 0.0113 & 0.4563 $\pm$ 0.0219 & 0.1169 \\ \bottomrule
    \end{tabular}
    \vspace{0.2cm}
    \caption{Hypervolume (HV) results based on cross-entropy loss}
    \label{tab:results_loss}
\end{table}

\begin{table}[!htbp]
    \scriptsize
    \centering
    \begin{tabular}{lcccccc} \toprule
\multirow{2}{*}{\bf Method}  & \multicolumn{2}{c}{\bf Multi MNIST} & \multicolumn{2}{c}{\bf Multi Fashion} & \multicolumn{2}{c}{\bf multi fashion+mnist} \\ \cmidrule(lr){2-3} \cmidrule(lr){4-5} \cmidrule(lr){6-7}
             &  TL  & BR & TL  & BR & TL  & BR  \\ \midrule
Single Task & 0.1993 $\pm$ 0.0035 & 0.2476 $\pm$ 0.0055 & 0.4151 $\pm$ 0.0059 & 0.4405 $\pm$ 0.0047 & 0.1219 $\pm$ 0.0050 & 0.3473 $\pm$ 0.0067 \\
Uniform & 0.2174 $\pm$ 0.0139 & 0.2621 $\pm$ 0.0123 & 0.4318 $\pm$ 0.0190 & 0.4523 $\pm$ 0.0186 & 0.1634 $\pm$ 0.0122 & 0.3849 $\pm$ 0.0093 \\
MGDA & 0.2189 $\pm$ 0.0105 & 0.2589 $\pm$ 0.0114 & 0.4316 $\pm$ 0.0079 & 0.4426 $\pm$ 0.0071 & 0.1558 $\pm$ 0.0070 & 0.3776 $\pm$ 0.0087 \\ \midrule
ParetoMTL & 0.2476 $\pm$ 0.0091 & 0.2706 $\pm$ 0.0062 & 0.4489 $\pm$ 0.0190 & 0.4380 $\pm$ 0.0155 & 0.2375 $\pm$ 0.0127 & 0.3644 $\pm$ 0.0066 \\
PHN$^*$ & 0.2347 $\pm$ 0.0056 & 0.2895 $\pm$ 0.0079 & 0.4761 $\pm$ 0.0273 & 0.4989 $\pm$ 0.0239 & 0.1871 $\pm$ 0.0106 & 0.4172 $\pm$ 0.0104 \\
PHN & 0.2308 $\pm$ 0.0074 & 0.2799 $\pm$ 0.0084 & 0.4761 $\pm$ 0.0273 & 0.4989 $\pm$ 0.0239 & 0.1681 $\pm$ 0.0102 & 0.4105 $\pm$ 0.0075 \\
COSMOS$^*$ & 0.2208 $\pm$ 0.0065 & 0.2592 $\pm$ 0.0076 & 0.4360 $\pm$ 0.0144 & 0.4540 $\pm$ 0.0176 & 0.2101 $\pm$ 0.0040 & 0.3656 $\pm$ 0.0063 \\
COSMOS & 0.2342 $\pm$ 0.0073 & 0.2684 $\pm$ 0.0080 & 0.4306 $\pm$ 0.0096 & 0.4453 $\pm$ 0.0127 & 0.3504 $\pm$ 0.0160 & 0.4289 $\pm$ 0.0124 \\ \bottomrule
    \end{tabular}
    \vspace{0.2cm}
    \caption{Results for each task in terms of cross-entropy loss}
    \label{tab:results_individual_loss}
\end{table}

\end{document}